\documentclass{article}

\usepackage{arxiv}

\usepackage[utf8]{inputenc} 
\usepackage[T1]{fontenc}    
\usepackage{hyperref}       
\usepackage{url}            
\usepackage{booktabs}       
\usepackage{amsfonts}       
\usepackage{nicefrac}       
\usepackage{lipsum}		
\usepackage{graphicx}
\usepackage{wrapfig}
\usepackage{doi}
\usepackage{bm}
\usepackage{color}
\usepackage{caption}
\usepackage{subcaption}
\usepackage{mathrsfs}
\usepackage{stmaryrd}
\usepackage{amsmath}
\usepackage{makecell}
\usepackage{float}
\usepackage{soul}
\usepackage{comment}

\title{Explainable Graph Pyramid Autoformer for \\ Long-Term Traffic Forecasting}

\author{ 
Weiheng Zhong\\
University of Illinois at Urbana-Champaign\\
Champaign, Illinois\\
\texttt{weiheng4@illinois.edu}
\And
Tanwi Mallick\\
Mathematics and Computer Science Division\\
Argonne National Laboratory, Lemont, IL\\
\texttt{tmallick@anl.gov}\\
\AND
Hadi Meidani \\
Department of Civil and Environmental Engineering\\
University of Illinois at Urbana-Champaign\\
Champaign, Illinois \\
\texttt{meidani@illinois.edu} \\
\AND
Jane Macfarlane\\
Sustainable Energy Systems Group\\
Lawrence Berkeley National Laboratory, Berkeley, CA \\
\texttt{jfmacfarlane@lbl.gov} \\
\AND
Prasanna Balaprakash\\
Mathematics and Computer Science Division \& Argonne Leadership Computing Facility\\
Argonne National Laboratory, Lemont, IL \\
\texttt{pbalapra@anl.gov}\\
}

\renewcommand{\shorttitle}{\textit{arXiv} Template}

\begin{document}
\maketitle

\begin{abstract}
Accurate traffic forecasting is vital to an intelligent transportation system. Although many deep learning models have achieved state-of-art performance for short-term traffic forecasting of up to 1 hour, long-term traffic forecasting that spans multiple hours remains a major challenge. Moreover, most of the existing deep learning traffic forecasting models are black box, presenting additional challenges related to explainability and interpretability. We develop Graph Pyramid Autoformer (X-GPA), an explainable attention-based spatial-temporal graph neural network that uses a novel pyramid autocorrelation attention mechanism.  It enables learning from long temporal sequences on graphs and improves long-term traffic forecasting accuracy. Our model can achieve up to 35\% better long-term traffic forecast accuracy than that of several state-of-the-art methods. The attention-based scores from the X-GPA model provide spatial and temporal explanations based on the traffic dynamics, which change for normal vs. peak-hour traffic and weekday vs. weekend traffic.
\end{abstract}


\section{Introduction}
\label{sec:Intro}


The most-congested 25 cities in the United States are estimated to have an annual congestion economic loss of \$50 billion  \cite{pishue2017us}. Intelligent transportation systems have the potential to reduce the economic losses incurred by growing transportation challenges. Accurate traffic forecasting is one of the key pillars of the intelligent transportation system. Although traffic forecasting is difficult because of the complexity of spatial and temporal dependencies, many deep learning models have achieved superior results for short-term traffic forecasts of up to 1 hour \cite{li2017diffusion, yu2017spatio, guo2019attention}. As the urban population rises, however, traffic dynamics and congestion usually last for hours. Hence, short-term forecasting alone cannot provide sufficiently rich information for proactive traffic management strategies, such as time signal control, to avoid potential traffic congestion \cite{yu2021long}. Modeling long-term traffic temporal dependency is difficult, however, since multiple temporal patterns entangle the long-term traffic dynamics. 

Since intelligent transportation systems that  use deep learning methods will affect everyone's life and governments' infrastructure investments, forecasting results must be trustworthy. Although most deep learning models have promising performance, they are labeled as ``black box,'' whose predictions cannot be explained. 
For establishing trustworthiness, explainable artificial intelligence (XAI) methods are crucial. Moreover, traffic dynamics are highly nonlinear and can have complex spatial dependencies and temporal patterns. XAI methods can help traffic managers  understand these complex traffic dynamics and bottlenecks, which are otherwise impossible to model analytically. 

We develop Explainable Graph Pyramid Autoformer (X-GPA), a new spatial-temporal graph neural network (GNN) model. Our approach is based on transformers, which have achieved high performances for long-sequence time series forecasting based on a self-attention mechanism \cite{https://doi.org/10.48550/arxiv.1706.03762}. 
To learn the temporal patterns for long-term traffic forecasting, we design a novel attention mechanism called pyramid autocorrelation attention. It extracts the temporal features hierarchically from time series using patch attention and uses an autocorrelation attention mechanism to combine the hierarchical features.   
To model the spatial dependencies, we adopt a graph attention layer \cite{https://doi.org/10.48550/arxiv.1710.10903} to aggregate node features based on (driving) distances and the pairwise relationship between two locations. With the new temporal learning together with the spatial dependency modeling, our X-GPA model achieves a high forecasting accuracy for long-term traffic forecasting and simultaneously explains the predictions with  attention scores. 
To that end, contributions of the paper are as follows:  
\begin{itemize}

    \item We propose a novel pyramid autocorrelation attention mechanism to overcome the computational complexity associated with learning long time series in spatial-temporal GNNs. 
    \item Our proposed X-GPA method defines a new state-of-the-art method for long-term traffic forecasting. Specifically, the X-GPA method achieves up to 35\% improvement in accuracy over the state-of-the-art spatial-temporal GNNs developed for traffic forecasting.
    \item Our X-GPA is the first ante hoc XAI method for long-term traffic forecasting, which provides attention-score-based spatial and temporal explanations for predictions. 
\end{itemize}

\section{Related work \label{sec.related_works}}

Here we  review the  methods with respect to traffic forecasting, XAI for traffic forecasting, and long-term time series forecasting, and we highlight the key differences of  the methods. 

\paragraph{Traffic forecasting} Early data-driven methods for traffic forecasting include historical average (HA) \cite{ermagun2018spatiotemporal} 
and ARIMA \cite{box2015time} 
In recent years, deep neural networks such as recurrent networks (RNNs) \cite{wen2017multi} and their variants (e.g., long short-term memory networks \cite{lai2018modeling}) have achieved superior performance. 
However, these methods do not leverage spatial dependencies of the traffic network. 
Recent works have shown that traffic forecasting requires both spatial and temporal dependencies. To that end, methods that combine GNNs with time series modeling have become state-of-the-art traffic forecasting methods. Diffusion-convolution recurrent neural network (DCRNN) \cite{li2017diffusion} used the diffusion convolution layer and gated recurrent layers to model the spatial  dependencies and temporal correlations, respectively. Spatio-Temporal Graph Convolutional Networks (STGCN) \cite{yu2017spatio} combined graph and temporal convolution. Attention Based Spatial-Temporal Graph Convolutional Networks (ASTGCN) \cite{guo2019attention} introduced a self-attention mechanism into STGCN for performance improvement. Nevertheless,  although these models have achieved state-of-art performances for short-term traffic forecasting (up to one hour), they are not competent for long-term traffic forecasting because they cannot learn patterns from long sequences or sequences having high computational complexity. We show that our X-GPA approach with the new pyramid autocorrelation attention mechanism obtains accuracy values that are significantly better than those  of existing spatial-temporal GNNs for long-term traffic forecasting.   
Recently Yu et al.~\cite{yu2021long} proposed a novel graph neural network for long-term traffic prediction using manually curated historical data as the input sequence, thereby reducing the complexity of learning from long sequences. In contrast, our X-GPA method takes the last seven days as the  input sequence and uses the attention mechanism to automatically capture the critical historical information, thereby reducing the manual effort required. 


\paragraph{Explainable traffic forecasting}
To overcome the challenges of black box models, some researchers  have started  exploring explainable methods to predict the traffic dynamics. Lai et al.~\cite{li2021multistep} adopted dynamic graph convolution to simulate the dynamics of the traffic system for short-term traffic prediction, which was considered as a partially explainable model. Cui et  al.~\cite{cui2020graph} developed graph Markov processes to predict  short-term traffic conditions under the setting of missing data. The parameters of the model after training are used to explain the traffic dynamics. However, all  these spatial temporal models are post hoc XAI methods, where the explanation extracted from the model is based on other numerical techniques. Moreover, most post hoc techniques are used under specific assumptions, which make the interpretation of the models untrustworthy \cite{vale2022explainable}. Hence these models cannot obtain an explanation for each prediction in the test dataset. 
To the  best of  our knowledge, this paper presents the  first use of  the attention mechanism to develop an interpretable GNN model for traffic forecasting. The results show that our model can automatically detect important historical features for each prediction and simulate the periodic temporal pattern of the dynamical systems. 


\subsection{Long-term time series forecasting}

Many traditional time series prediction models fail to capture the long-term temporal dependencies and thus either accumulate extremely high errors or suffer from huge computation cost \cite{ermagun2018spatiotemporal, box2015time, wen2017multi}. In recent years, the attention mechanism has become the key component in neural networks for long-term time series forecasting \cite{https://doi.org/10.48550/arxiv.1706.03762, cirstea2022triformer}. 
Among all the models using the attention mechanism, the transformer \cite{https://doi.org/10.48550/arxiv.1706.03762} using the self-attention mechanism has obtained state-of-the-art performance in time series data modeling with $O(L^2)$ computational complexity, where $L$ is the input sequence length of the time series. 
The high computation complexity of the self-attention mechanism \cite{https://doi.org/10.48550/arxiv.1706.03762} is, however, an obstacle to applying the model for information extraction from long-sequence input data. Therefore, many works adapt the architecture of transformers for higher efficiency. Reformer \cite{kitaev2020reformer} reduced the complexity to $O(L \log L)$ using the local-sensitive hashing attention and reversible residual layers. Informer \cite{zhou2021informer} also achieved  computation complexity of $O(L \log L)$ with a sparse self-attention mechanism. Pyraformer \cite{liu2021pyraformer} explored the multiresolution representation of the time series based on the pyramidal attention module;  its time and space complexity scale linearly with time length $L$. Triformer \cite{cirstea2022triformer} also achieved linear complexity of $O(L)$ by developing a novel attention mechanism called patch attention.
Furthermore, some transformer variants were developed for better prediction performance by applying the attention mechanism in the frequency domain. Autoformer \cite{https://doi.org/10.48550/arxiv.2106.13008} proposed a novel autocorrelation  attention mechanism by using a fast Fourier transform, which achieved significant improvement compared with  non-frequency-based transformers. 

Built on Autoformer and Triformer, our model adopts a new pyramid autocorrelation mechanism by combining patch attention and autocorrelation attention, resulting in computation complexity of $O(L)$ while maintaining the same level of information utilization capability. Moreover, the existing long-term forecasting methods do not leverage spatial dependencies, which are crucial for traffic forecasting. Equipped with the new pyramid autocorrelation mechanism and graph attention layer, our proposed X-GPA achieves significantly better forecasting accuracy than the state-of-the-art Autoformer for long-term forecasting.

\section{Explainable Graph Pyramid Autoformer for
Long-Term Traffic Forecasting \label{sec:Med}}


We consider the most commonly adopted traffic forecasting setup \cite{li2017diffusion, yu2017spatio, guo2019attention}, wherein a traffic network is modeled as an undirected graph $G=(V, E, A)$, where $V$ is a set of all vertices representing (sensor)  locations  $E$ is a set of all edges representing the connection between the locations, and $A \in \mathbb{R}^{N \times N}$ is the the adjacency matrix representing the connectivity among nodes, where $N=|V|$ is the number of vertices. The sensors in the vertices collect measurements of the system state with a fixed sampling frequency. The dimensionality of the system features that each sensor collects is $D$. The collected feature sequence $\bm{\chi}_{t-L+1}^t = \{ x^{t-L+1}, x^{t-L+2},...,x^t \}$ , where $x_t \in \mathbb{R}^{N \times D}$ denotes the feature matrix of the graph observed at time $t$. The prediction task is to find a mapping function $\it f$ from the previously observed feature matrix to the future feature matrix:
$
  \bm{\chi}_{t+1}^{t+Q} = \it{f} ( \bm{\chi}_{t-L+1}^t; A ).
$

Instead of fitting a model to simulate the traffic evolution \cite{li2017diffusion, ermagun2018spatiotemporal}, our model extracts the useful information from the input sequence based on the attention mechanism and predicts the future based the spatial-temporal pattern of the history input sequence. 


\subsection{Pyramid autocorrelation attention for learning temporal patterns \label{subsec.ac_atten}}

The pyramid autocorrelation attention mechanism that we propose consists of two parts: patch attention to derive hierarchical representations of the time series and an autocorrelation attention mechanism to learn the periodic pattern of each representation. 
In patch attention, the time series is divided into small patches, and the attentions of hidden states of the timestamps in each patch are computed to derive a hidden state of single pseudo timestamp (this will look like inverted pyramid shape). 
In the same way, patch attentions are applied on the derived pseudo time steps hierarchically. Consequently, at each level of the hierarchy, temporal features are aggregated at different time resolutions (for example, 5, 15, and 30 minutes). Then the autocorrelation attention is applied for each level of the hierarchy to extract periodic information and patterns from the pseudo time series. 

\subsubsection{Patch attention}
We build on the patch attention operator introduced in Triformer \cite{https://doi.org/10.48550/arxiv.2204.13767}, wherein the patch attention is used to reduce the computational complexity of transformer for long input sequences. We use patch attention for reducing computation costs and extracting multiscale traffic time series patterns.


For a given the input time series  ${X}$ with time length $L$, applying patch attention of size $ps$ will result in a pseudo time series ${Y}$ of time length $L/{ps}$. This is achieved as follows. The input time series of length $L$ is divided into $L/{ps}$ patches in the temporal direction. The $j$th patch is denoted as $P_j = \{ X_{(j-1) \cdot {ps}+1}, ..., X_{j \cdot {ps}}\}$, and $Y_j$ is the output of the $j$th patch.
The attention score $S_q^{p_j}$ of the $q$th value in the  $j$th patch is calculated in two steps. First the keys for all the values in a patch except for the $q$th value are computed:
\[
  Keys = \bm{\bigparallel}_{i \in P_j, i \neq q}^{ps} F_K(X_i; \theta_K), 
\]
where $\bm{\bigparallel}$ is the sum of concatenation and $F_K(\cdot)$ is the key mapping with the trainable parameters $\theta_K$. Then the attention score $ S_q^{p_j}$ of the $q$th value is calculated as follows:
\[
  S_q^{p_j} = \sigma(W_{patch} [F_Q(X_q; \theta_Q) | Keys],
\]
where $|$ is the concatenation  operation,  $F_Q(\cdot)$ is the query mapping with parameters $\theta_Q$, $W_{patch}$ is a set of trainable parameters, and $\sigma$ is the nonlinear activation function. 
Then, a softmax function is applied to aggregate the feature with the weights of attention scores as follows:
\begin{align*}
    S_1^{'p_j}, ..., S_{ps}^{'p_j} &= \rm{softmax} \{ S_1^{p_j}, ..., S_{ps}^{p_j} \}, \\
    Y_j &= \sum_q^{ps} S_q^{'p_j} \cdot F_V(X_i; \theta_V),
\end{align*}
where $S_q^{'p_j}$ is the normalized $q$th attention score in $j$th patch and $F_V$ represents the value mapping. 

\subsubsection{Autocorrelation attention}

To extract the periodic temporal pattern of the input sequence, we use an autocorrelation attention mechanism \cite{https://doi.org/10.48550/arxiv.2106.13008}. For a real-valued discrete time series $\{ \bm{\chi_t} \}$ with time length $L$, we can obtain the autocorrelation $R_{\chi \chi} (\tau)$ as a function of the time shift $\tau$ efficiently by using fast Fourier transforms (FFTs) based on the Wiener–-Khinchin theorem \cite{wiener1930generalized}:
\[
  R_{\chi \chi} (\tau) = \mathscr{F}^{-1} ( \mathscr{F} (\chi_t) \mathscr{F}^{conj} (\chi_t) ),
\]
where $\tau \in \{1,...,T\}$ is the delay time length,  $\mathscr{F}$ denotes the FFT, $\mathscr{F}^{-1}$ is its inverse, and $conj$ represents the conjugate operation. The normalized $ \frac{R_{\chi \chi} (\tau_i)}{\sum_k R_{\chi \chi} (\tau_k)}$ can be considered as a possibility for the time series to repeat itself with the time shift $\tau_i$. Hence, we consider the future traffic prediction as the linear combination of the time-shifted historical data based on this possibility. Applying this concept to the attention mechanism, we can calculate the  autocorrelation by the keys and queries:
\[
  R_{Q_t, K_t} (\tau) = \mathscr{F}^{-1} ( \mathscr{F} (Q_t) \mathscr{F}^{conj} (K_t) ),
\]
where $Q_t$ and $K_t$ respectively are the time series after applying query mapping and  key mapping on the input time series $\chi_t$. Following the implementation of \cite{https://doi.org/10.48550/arxiv.2106.13008}, we choose the most possible $k$ delay values to aggregate the temporal features for the purpose of reducing the memory complexity:
\[
  \tau_1, ..., \tau_k = \rm{arg_{\tau} Topk} \{ R_{Q_t, K_t} (\tau), \}
\]
where $\rm{arg_{\tau} Topk}(\cdot)$ gets the indices of the $k$ largest autocorrelations. Then we normalize the autocorrelation as our temporal attention scores using the softmax function \cite{https://doi.org/10.48550/arxiv.1706.03762}:
\[
  S_1, ..., S_k = \rm{softmax} \{ R_{Q_t, K_t} (\tau_1), ..., R_{Q_t, K_t} (\tau_k). \}
\]
Now we can aggregate the temporal feature to calculate the output feature $\chi^{out}$ based on the scores of the autocorrelation:
\[
  \chi^{out} = \sum_i^k \rm{Roll} (V_t, \tau_i) * S_i,
\]
where $V_t = [v^1, ..., v^t]$ is the time series after applying value mapping on the input time series $\chi_t$  and where  $\rm{Roll} (V_t,  \tau)$ represents the rolling operation to $V_t$ in temporal dimension with time delay $\tau$:
\[
  \rm{Roll} ([v^1, ..., v^t], \tau) = [v^{\tau+1}, ..., v^t, v^1, ...,v^{\tau} ].
\]

\begin{figure}[ht]
    \centering
    \includegraphics[width=10cm]{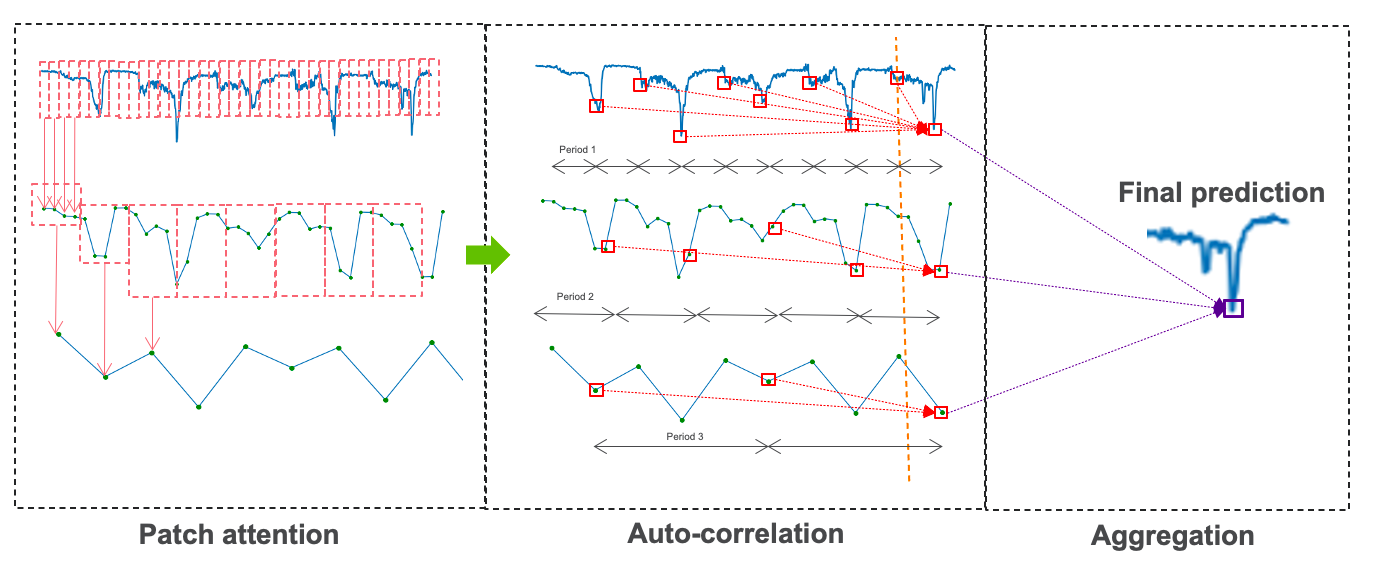}
    \caption{\footnotesize Description of pyramid autocorrelation attention mechanism. The lower-dimensional representations of the input time series are first derived by using patch attention. Then we extract the periodic temporal pattern of each representation based on autocorrelation attention.}
    \label{patch_atten}
\end{figure}

After applying multiple-patch attention layers, we can derive multiple shorter sequence representations $\{Y^1, Y^2, ...\}$. We apply autocorrelation attention to each time series to calculate the hidden state of the predicted time steps. The architecture of the pyramid autocorrelation attention mechanism is shown in Figure \ref{patch_atten}.

\subsection{Graph attention network for learning spatial dependency\label{subsec:GAL}}

To model the spatial dependency among the nodes in a graph, we adopt graph attention networks \cite{https://doi.org/10.48550/arxiv.1710.10903}. 
The input is a set of node features $H^t = \{ h_1, h_2, ..., h_N \}, h_i \in \mathbb{R}^{D}$ at a specific time $t$, where $N$ is the number of nodes.  The graph attention layer aggregates the neighbor nodes' features of a given node to produce new node features $H'^t = \{ h'_1, h'_2, ..., h'_N \}, h'_i \in \mathbb{R}^{D}$. We first compute the importance of node $j$ to node $i$ $I_{ij}$ of $h_j$ to $h_i$ by
$
  I_{ij} = \sigma(W_{sp} [ F_Q(h_i; \theta_Q) | F_K(h_j; \theta_K) || d_{ij} ]),
$
where $F_Q(\cdot) \in \mathbb{R}^D \shortrightarrow \mathbb{R}^D$ is the query mapping and $F_K(\cdot) \in \mathbb{R}^D \shortrightarrow \mathbb{R}^D$ is the key mapping with the trainable parameters $\theta_Q$ and $\theta_K$, respectively. $W_{sp} \in \mathbb{R}^{{2D+1}}$ is a set of trainable parameters, $d_{ij}$ represents the distance between node $i$ and node $j$, $|$ is the concatenation operation, and $\sigma$ is the nonlinear  activation function. Then we inject graph structure into the attention mechanism by performing masked attention. The attention coefficient $\alpha_{ij}$ from $h_j$ to $h_i$ is computed as follows:
$
  \alpha_{ij} = \frac{exp(I_{ij})} {\sum_{k \in \mathbb{N}_i} exp(I_{ik})},
$
where $\mathbb{N}_i$ represents the neighbor nodes of  node $v_i$. Then the new node feature is computed as follows:
$
  h'_i = W_2 \sigma ( \sum_{j \in \mathbb{N}_i} \alpha_{ij} F_V(h_j; \theta_V)),
$
where $F_V \in \mathbb{R}^D \shortrightarrow \mathbb{R}^D$ is the value mapping of parameters $\theta_V$ and $W_2 \in \mathbb{R}^{D \times D}$ is a set of trainable parameters. 

\subsection{Model architecture}

\begin{figure}[ht]
    \centering
    \includegraphics[width=10cm]{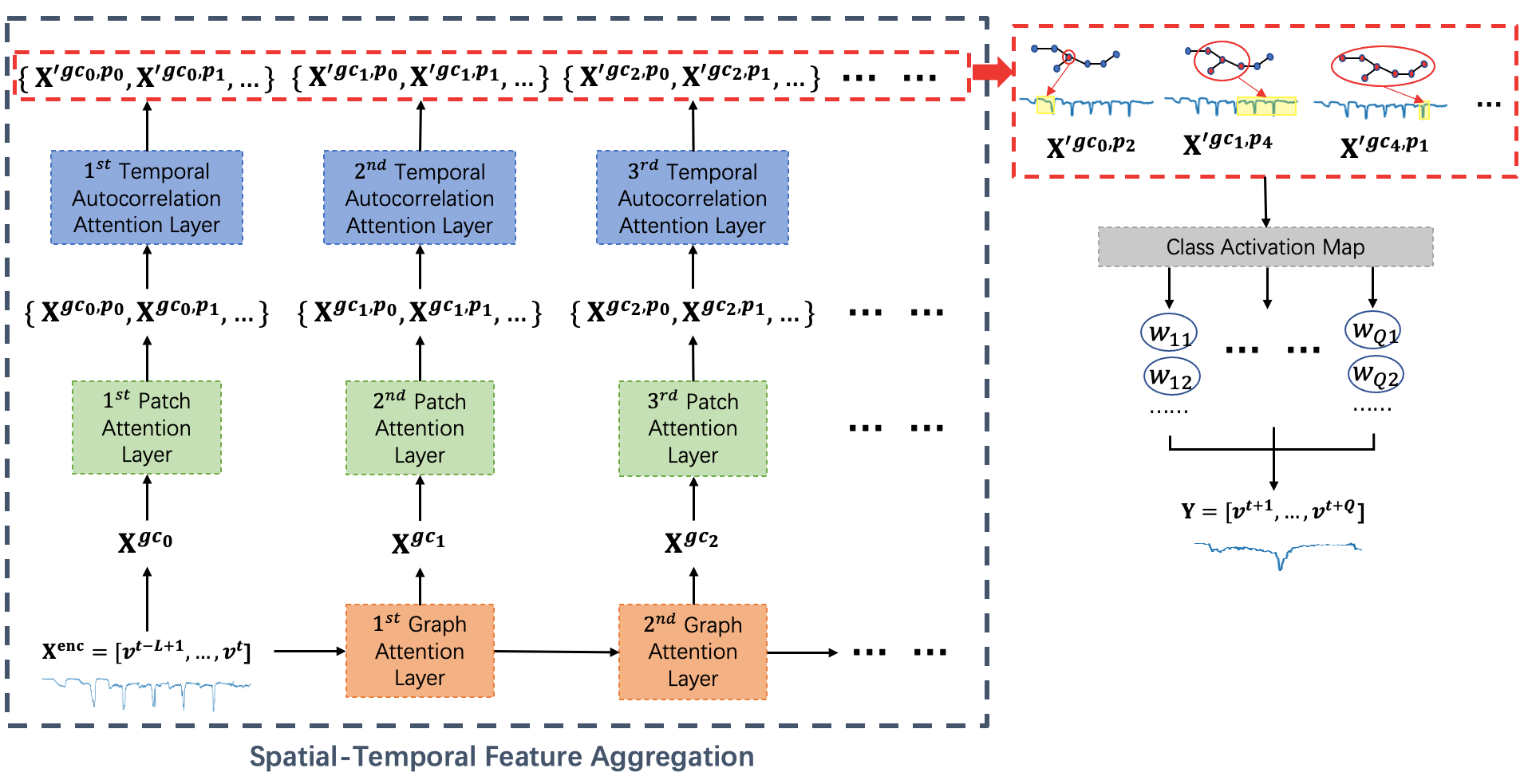}
    \caption{\footnotesize Framework of graph pyramid autoformer.}
    \label{model_framework}
\end{figure}

Our proposed X-GPA model architecture is shown in Figure \ref{model_framework}. The model can be divided into two parts. In the module of spatial-temporal feature aggregation, we first apply a graph attention layer \ref{subsec:GAL} to aggregate the feature of the neighbor nodes. The output $\chi^{gc_i}$ represents the aggregated features after $i$ graph attention layers. Then we apply temporal patch attention to aggregate traffic features in the temporal dimension. The output $\chi^{p_j}$ represents the aggregated features after $j$ patch attention layers. After applying the spatial graph attention layers and temporal patch attention layers, we obtain a set of spatial-temporal aggregated traffic feature $\{\chi^{gc_i,p_j}\}_{i=0,j=0}^{M_{gc}, M_{p}}$, where $M_{gc}$ is the number of graph attention layers and $M_{p}$ is the number of patch attention layers.
Using the autocorrelation mechanism, we derive the hidden features of the predicted time slots $\chi'$ by aggregating the historical traffic data based on the periodic pattern of the time series. All selected features are used for the prediction of each future timestep. We then utilize class activation mapping \cite{https://doi.org/10.48550/arxiv.1512.04150} to help  derive the weights of all feature maps $\{w_{i,j}\}_{i=0,j=0}^{M_{gc}, M_{p}}$, where $w_{i,j}$ is the weight for the feature map $\chi^{gc_i,p_j}$. This is given by
$
  \{w_{i,j}\}_{i=0,j=0}^{M_{gc}, M_{p}} = \exp[F(\{\chi'^{gc_i,p_j}\}_{i=0,j=0}^{M_{gc},M_{p}}; \bm{\theta})],
$
where $F$ is a neural network model parameterized by $\bm{\theta}$ and $\exp$ is the exponential operation to ensure that the weights are positive. The hidden feature of the predicted traffic data is calculated by
$
  v^{t+j} = \sum_{i,j} w_{i,j} \chi'^{gc_i,p_j}.
$

\section{Explanation extraction}

\begin{figure}[ht]
    \centering
     \includegraphics[width=10cm]{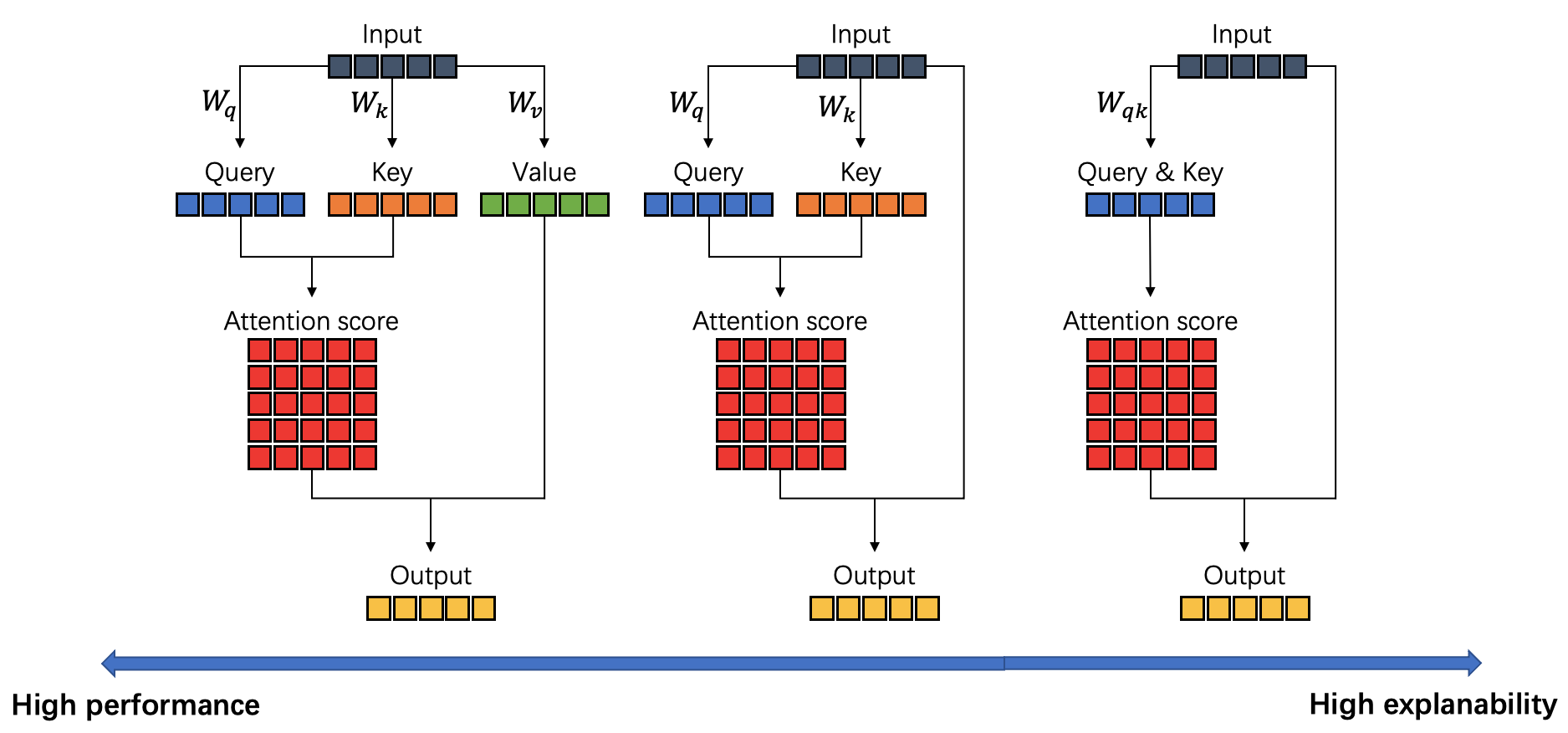}
     \label{sensor_loc}
    \caption{\footnotesize Description of different attention mechanisms.}
    \label{explainable_attn}
\end{figure}

The attention mechanism is a popular approach to explain the deep learning model predictions. In a traditional self-attention layer, the output is considered as a linear combination of input features. To increase the explainability of the model, we develop two variants of the single-headed attention mechanism by either removing the value mapping or removing values mapping and query mapping simultaneously, as shown in Figure \ref{explainable_attn}. In the first variant, we define value mapping as identical mapping $F_V(x) = x$. In the second variant, we adopt the same function as our query mapping and key mapping $F_Q = F_K$ while using the identical mapping as the value mapping at the same time. When using these two attention mechanisms, even though we stack multiple graph attention layers and temporal attention layers, the final output becomes a linear combination of the original input. In our model, we tested three different kinds of attention mechanism and adopted the most explainable one.

Recall that in the graph attention layer we define $\alpha_{kl}$ as the importance of node $l$ to node $k$, and in the temporal attention layer we define $S_q$ as the attention score with respect to time delay $\tau_q$. We define $\mathbb{GC}$ as the operation of the graph attention layer and define $\mathbb{TC}$ as the operation of the temporal attention layer. 

The input of each attention layer is defined as $\mathbb{I} \in \mathbb{R}^{N \times T_1 \times D}$, and the output is defined as $\mathbb{O} \in \mathbb{R}^{N \times T_2 \times D}$, where $N$ is the number of nodes, $T_1, T_2$ is the number of time steps, and $D$ is the number of features. Hence we can derive the relationship between the input and the output of multiple graph attention layers as follows:
\[
  \mathbb{O}_{k,:,:} = \mathbb{GC} ( \sum_{l \in \mathbb{N}_k} \alpha_{kl} \mathbb{I}_{l,:,:}).
\]
Similarly, the relationship between the input and the output of multiple temporal attention layers is
\[
  \mathbb{O}_{:,p,:} = \mathbb{TC} (\sum_q S_q * \mathbb{I}_{:,p+\tau_q,:} ).
\]
Therefore, the prediction of node $k$ at time step $p$ can be formulated as
\[
  \chi^{out}_{k,p,:} = \mathbb{TC} [\sum_q S_q * \mathbb{GC} ( \sum_{l \in \mathbb{N}_k} \alpha_{kl} \chi_{l,:,:})_{:,p+\tau_q,:} ].
\]
Hence, the importance of the feature of node $l$ at time step $p+\tau_q$ to the prediction of the feature of node $k$ at time step $p$ is $S_q * \alpha_{kl}$. 

\section{Numerical experiments \label{sec:Traffic_Result}}


We used PeMS-BAY and Metr-LA, two real-world traffic datasets  \cite{li2017diffusion}, which have been used in a number of prior traffic forecasting studies \cite{li2017diffusion, yu2017spatio, guo2019attention}. The recorded traffic speeds were aggregated into 5-minute intervals. The PeMS-BAY dataset includes information  from  325 sensors in the San Francisco Bay area collecting five months of data from Jan. 1 to May 31, 2017. The Metr-LA dataset includes information from 207 sensors  in the highway of Los Angeles County from March 1  to June 30, 2012. 
We split the data into three parts: 15 weeks of traffic data for training (70\%), 2 weeks of traffic data for validation (10\%), and  4 weeks of traffic data for testing (20\%).

We used the following baselines to compare the performance of our proposed X-GPA: historical average \cite{ermagun2018spatiotemporal}, STGCN \cite{Yu_2018},  ASTGCN \cite{ye2022attention}, DCRNN \cite{li2017diffusion}, and Autoformer \cite{https://doi.org/10.48550/arxiv.2106.13008}.

We used mean absolute error (MAE) as the criterion to quantify the forecasting error of each time step $t$: $MAE_t =\frac{\sum_j^N |x^t_j - x^{pre,t}_j|}{N}$, 
where $N$ is the number of sensors, $x^t$ represents the ground truth traffic future data at time step $t$, and $x^{pre,t}$ represents the model forecast at time step $t$.

We conducted experiments under four settings:
Case 1: last 1-hour traffic as input time horizon; Case 2: last 1-hour traffic along with daily periodic data input time horizon (for example, to forecast 8--9 am traffic on Monday, the traffic conditions of 7--8 am on that Monday along with 8--9  am traffic of Monday to Sunday), which will be beneficial for model performance since traffic data has strong daily periodic pattern; Case 3: last 7 days as input time horizon, which will enable the models to leverage the strong weekly periodic pattern; and Case 4: last 7 days input time horizon but 12 hours forecast time horizon. Although X-GPA is designed for long-term forecasting, first we  compared our approach with several state-of-the-art short-term traffic forecasting methods (Cases 1--3). We found that our approach is either superior or comparable to the existing methods. 
We refer the reader to the Appendix \ref{Sec.app} for more detailed results of short-term traffic forecasting results.

\subsection{Long-term traffic forecasting}
Here we show that X-GPA completely outperforms previous state-of-the-art methods for long-term traffic forecasting.

The results for Case 4 are shown in Table \ref{long_term_speed_prediction_performance}. We used seven days of data as input and forecast future 12 hours of traffic data. Because of the extremely high computation cost of DCRNN and ASTGCN (see the Appendix), we did not include them in the comparison.

The results show that X-GPA achieves accuracy values that are better than the other methods for all the time horizons from 2 hours to 12 hours. HA and STGCN cannot achieve satisfactory accuracy for long-term traffic forecasts. Autoformer outperforms both HA and STGCN. However, X-GPA achieved much higher accuracy than the other three methods, with an MAE of only 2.86 for 12-hours-ahead prediction, a value that these three methods cannot achieve even for 2-hour forecasts. The trend is similar on the Metra-LA dataset as well. We observed that the overall accuracy improvement compared with the best baseline is over 35\%. 

\begin{table}[h]
\caption{MAE (mph) comparison for Case 4}
\centering
\begin{tabular}{c c c c c}
\hline
{} & \multicolumn{4}{c}{PEMS-BAY dataset}\\
Model & 2 hours & 4 hours & 8 hours & 12 hours \\
\hline
HA & 5.42 $\pm$ 0.00 & 5.42 $\pm$ 0.00 & 5.42 $\pm$ 0.00 & 5.42 $\pm$ 0.00  \\
STGCN & 5.09 $\pm$ 0.19 & 5.12 $\pm$ 0.21 & 5.17 $\pm$ 0.31 &  5.20 $\pm$ 0.40 \\
Autoformer & 4.05 $\pm$ 0.11 & 4.06 $\pm$ 0.15 & 4.08 $\pm$ 0.20 & 4.12 $\pm$ 0.25 \\
{X-GPA}  & \textbf{2.72 $\pm$ 0.08} & \textbf{2.77 $\pm$ 0.12} & \textbf{2.81 $\pm$ 0.16} & \textbf{2.86 $\pm$ 0.27} \\
\hline
\end{tabular}

\medskip

\begin{tabular}{c c c c c}
\hline
{} & \multicolumn{4}{c}{Metr-LA dataset}\\
Model & 2 hours & 4 hours & 8 hours & 12 hours \\
\hline
HA & 10.14 $\pm$ 0.00 & 10.14 $\pm$ 0.00 & 10.14 $\pm$ 0.00 & 10.14 $\pm$ 0.00  \\
STGCN & 9.02 $\pm$ 0.19 & 9.08 $\pm$ 0.21 & 9.11 $\pm$ 0.31 &  9.19 $\pm$ 0.40 \\
Autoformer & 7.15 $\pm$ 0.12 & 7.26 $\pm$ 0.15 & 7.48 $\pm$ 0.20 & 7.62 $\pm$ 0.27 \\
{X-GPA}  & \textbf{4.80 $\pm$ 0.15} & \textbf{4.87 $\pm$ 0.19} & \textbf{4.92 $\pm$ 0.25} & \textbf{4.98 $\pm$ 0.27} \\
\hline
\end{tabular}
\label{long_term_speed_prediction_performance}
\end{table}

\subsection{Explaining forecasts}

\subsubsection{Short-term forecasting}
\label{short_term_pred_interpret}

\begin{figure}[ht]
    \begin{subfigure}[b]{0.45\textwidth}
         \centering
         \includegraphics[width=7cm]{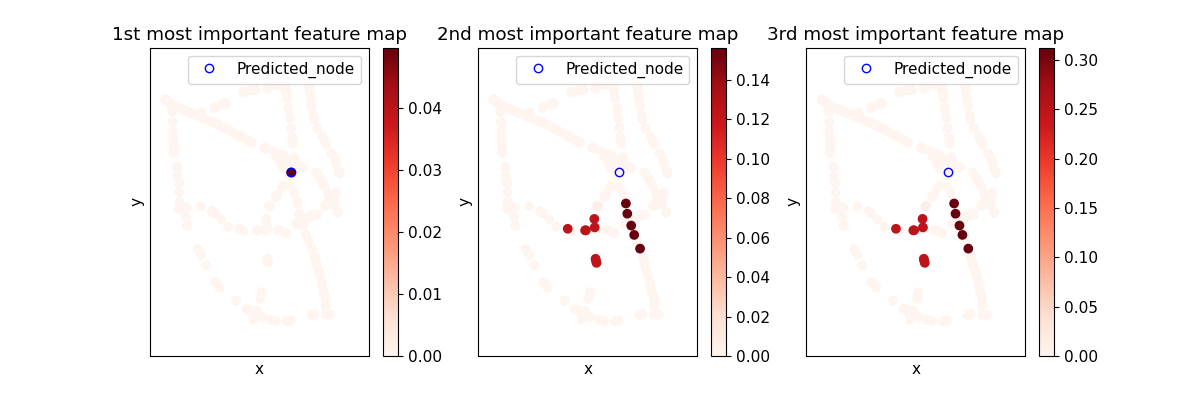}
         \caption{9:00am on Tuesday}
         \label{short_term_spatial_attn_peak}
     \end{subfigure}
     \begin{subfigure}[b]{0.45\textwidth}
         \centering
         \includegraphics[width=7cm]{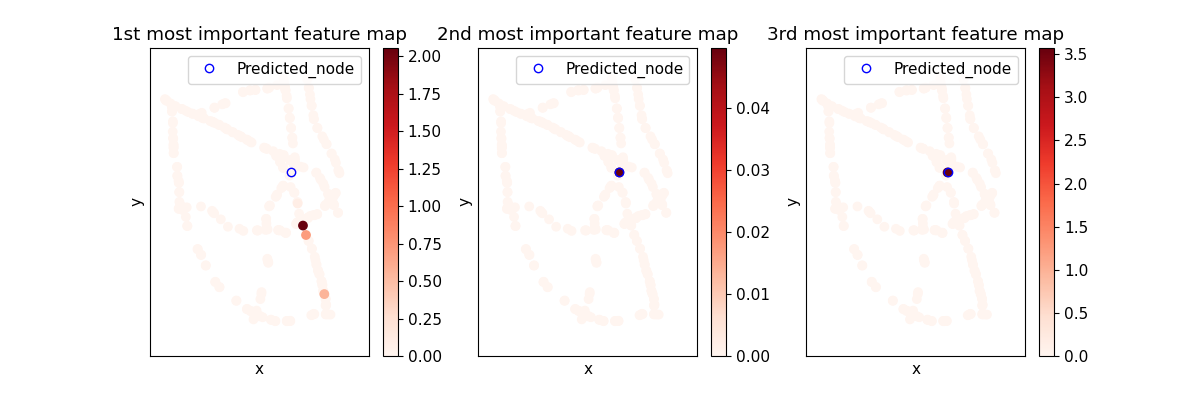}
         \caption{12:00pm on Sunday}
         \label{short_term_spatial_attn_non_peak}
     \end{subfigure}
    \caption{\footnotesize Spatial distribution of one-hour-ahead prediction for peak hours (top) and non-peak hours (bottom). Most attention accumulates on the target node itself for short-term prediction (within one hour).}
    \label{short_term_spatial_atten}
\end{figure}

In this section we  visualize the traffic segments that our model focuses on. To better show our model's ability to focus on important traffic features using the attention mechanism, we show additional results of the model attention scores of Case 3. We  carried out more detailed analysis only for the model of Case 3 because we provide more information (i.e., seven days of traffic data) to the model, which increases the difficulty of picking useful information. In Figure \ref{short_term_spatial_atten} we plot the spatial attention distribution of top three important feature maps for future one-hour traffic prediction. We observe that the model focuses on the historical data of the target node for non-peak-hour prediction; on the other hand, it pays attention to the neighbor nodes for peak-hour prediction. We also observe that the nodes near the 
interconnection of different highways  have a significant effect on our target node prediction, which is reasonable based on the dynamics of traffic movement. Moreover, we  observe that spatial attention is  distributed not just around the target node but along a specific path. One possible reason is that the model can capture the traffic movement pattern;  this still requires more detailed validation. 


\begin{figure}[ht]
    \centering
    \begin{subfigure}[b]{0.45\textwidth}
         \includegraphics[width=7cm]{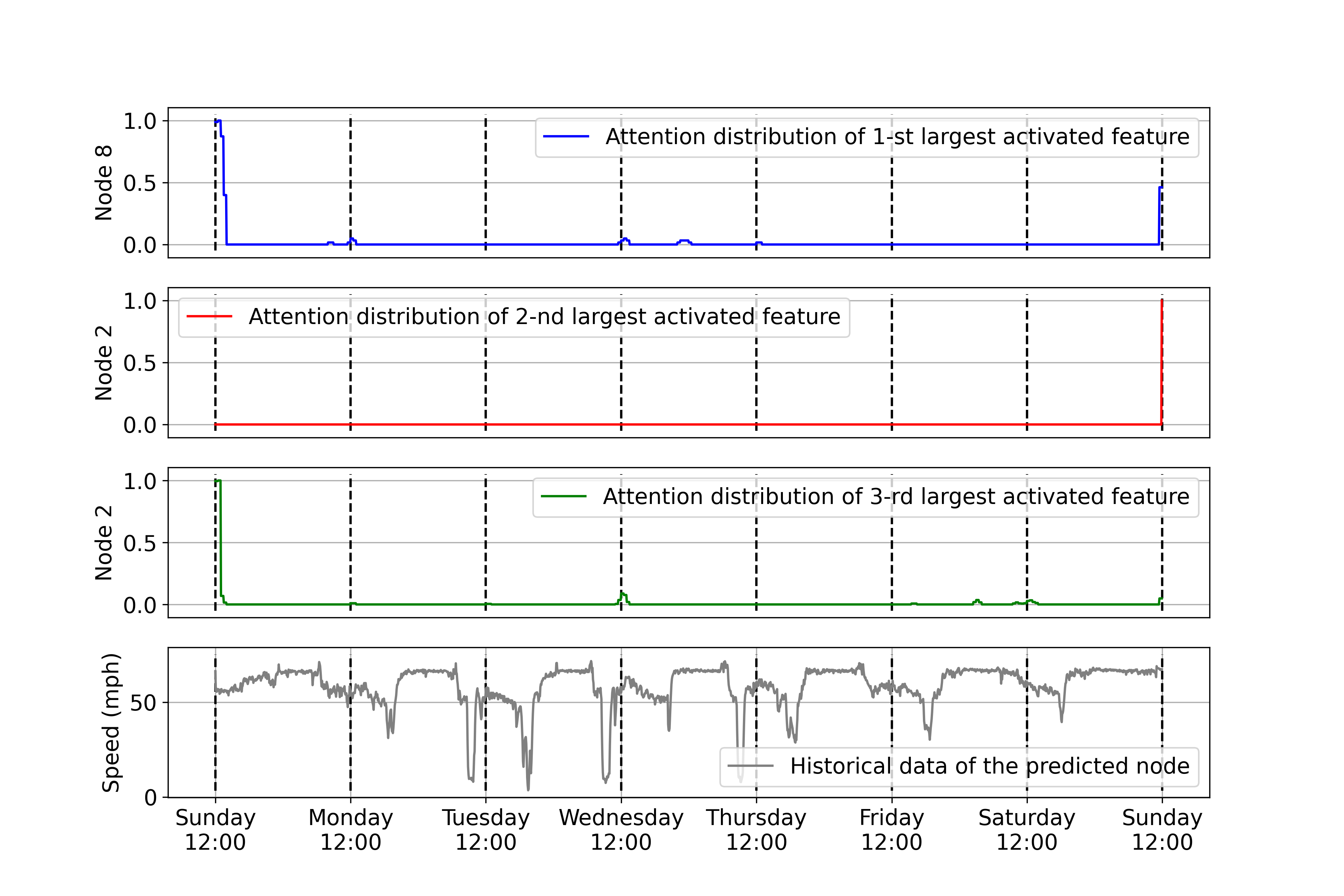}
         \caption{One-hour-ahead prediction for 12:00--1:00 pm on Sunday.}
         \label{non_peak_hour_attn_12}
     \end{subfigure}
     \hspace{0.5cm}
     \begin{subfigure}[b]{0.45\textwidth}
         \includegraphics[width=7cm]{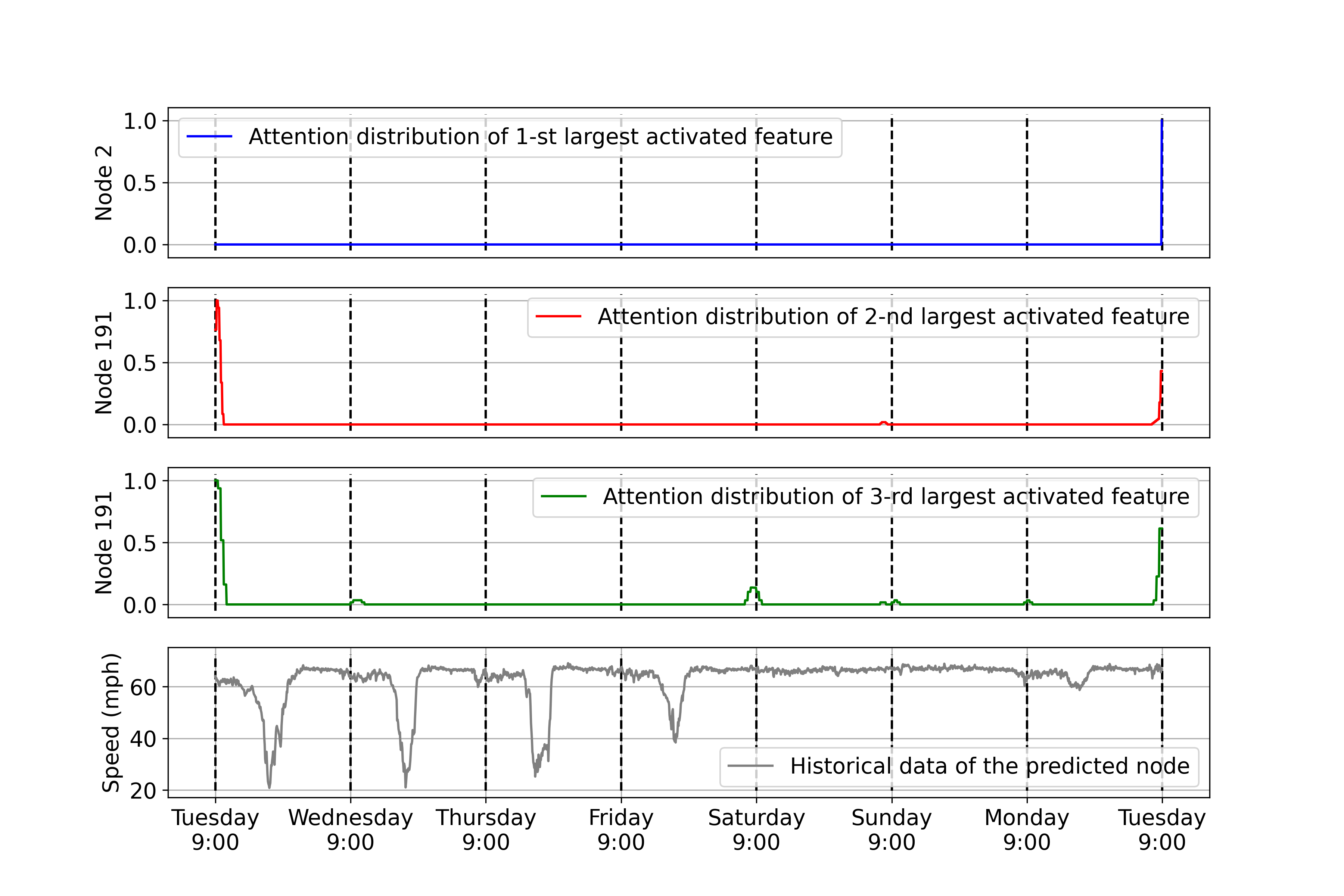}
         \caption{One-hour-ahead prediction for 9:00--10:00 am on Tuesday.}
         \label{peak_hour_attn_12}
     \end{subfigure}
    \caption{\footnotesize Comparison of attention distribution with the Pearson correlation for non-peak hours (top) and peak hours (bottom). Models pay attention to the most recent traffic and traffic data of the same time from the last week. Attention on the history will be higher for peak-hour traffic prediction.}
    \label{short_term_temporal_atten}
\end{figure}

For each feature map, we show the temporal attention distribution of the most influential node. The most influential node is defined as the node of highest spatial attention, which has deepest color in  Figure \ref{short_term_spatial_atten}. We show the temporal attention of the historical traffic data of the most influential nodes in Figure \ref{short_term_temporal_atten}. Based on the second and third feature maps of Figure \ref{non_peak_hour_attn_12}, most attention is accumulated on the same time from the last week and the most recent traffic data of the target node itself. For peak hour prediction, we see  that most attention accumulates on the recent traffic conditions of the target node history data. The most recent traffic and the traffic conditions of the same time from the last week are also important for prediction.

\subsubsection{Long-term forecasts}

\begin{figure}[ht]
    \centering
     \begin{subfigure}[b]{0.45\textwidth}
         \includegraphics[width=6cm]{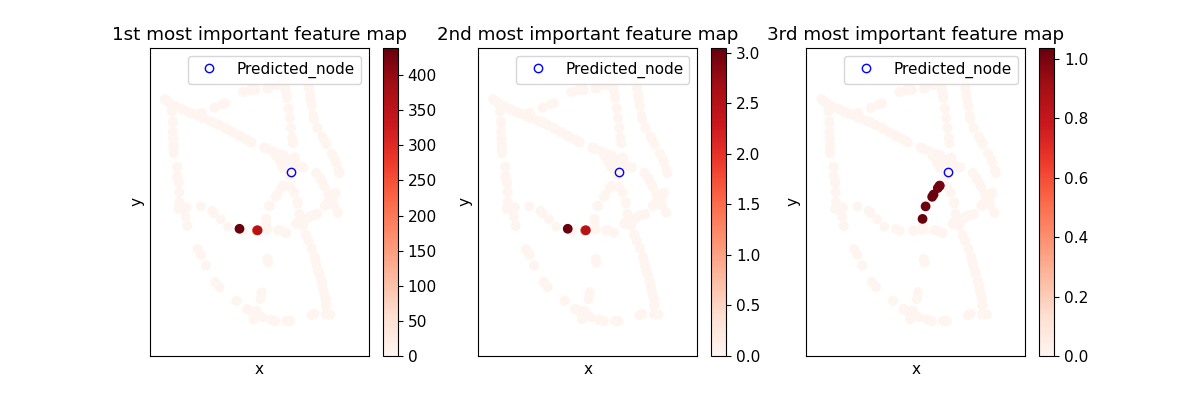}
         \caption{Prediction for 12:00   pm--00:00 am on Sunday}
         \label{long_term_spatial_non_peak}
     \end{subfigure}
     \hspace{0.5cm}
     \begin{subfigure}[b]{0.45\textwidth}
         \includegraphics[width=6cm]{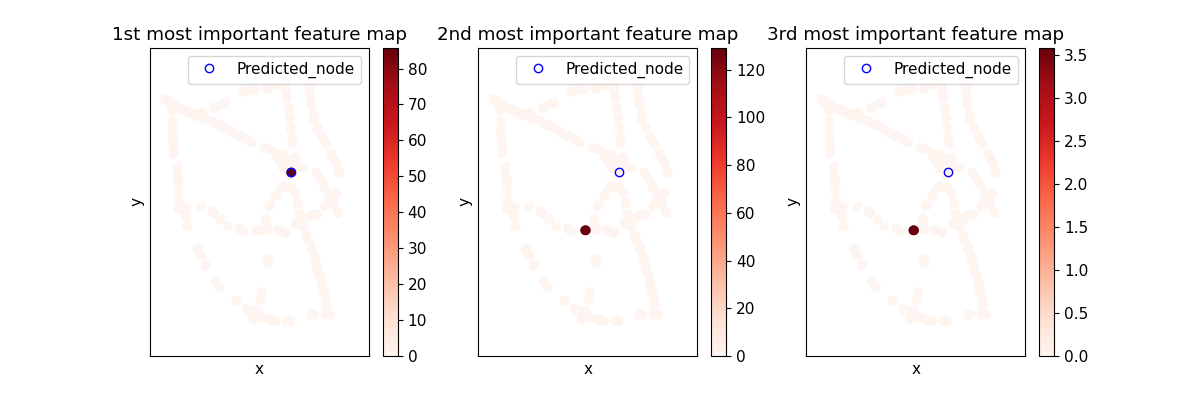}
         \caption{Prediction for 9:00 am -- 9:00 pm on Monday}
         \label{long_term_spatial_peak}
     \end{subfigure}
    \caption{\footnotesize Spatial attention distribution of the long-term prediction model for short-term traffic prediction (top) and long-term traffic prediction (bottom). The attention will shift from the target node to neighbor nodes for further future traffic prediction.}
    \label{long_term_spatial_atten}
\end{figure}

We observe that for long-term traffic forecasts, the model pays more attention to the neighbor nodes but not the target nodes themselves. We show the spatial attention for the long-term traffic prediction model (Case 4) in Figure \ref{long_term_spatial_atten}. For peak-hour traffic prediction, we observe that the history of the target traffic data is still important 
However, to predict non-peak hour traffic, our model does not use the information of the target node but relies more on neighbor nodes. This indicates that the spatial correlation is important for long-term prediction.

\begin{figure}[ht]
    \centering
    \begin{subfigure}[b]{0.45\textwidth}
         \includegraphics[width=7cm]{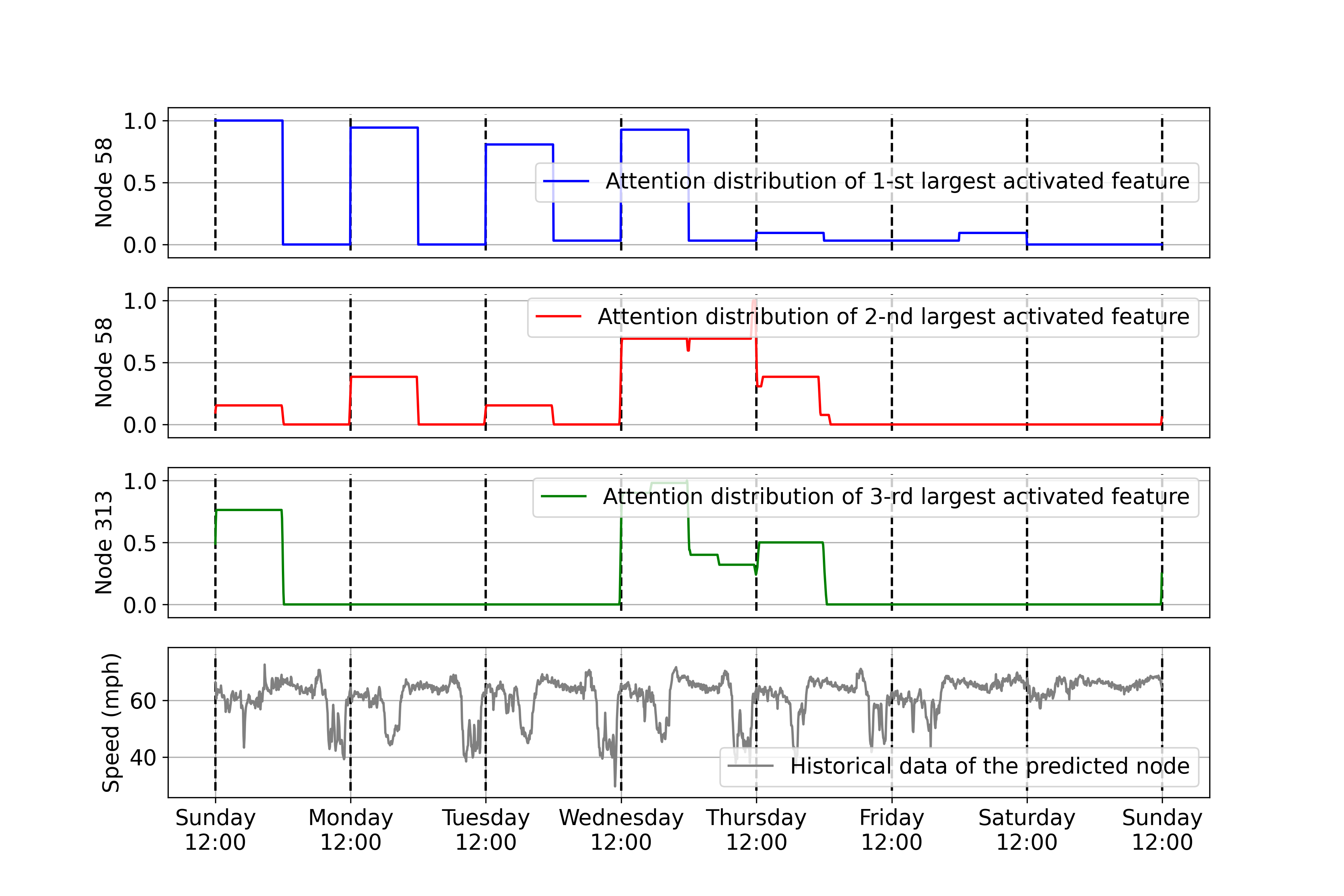}
         \caption{Prediction for 12:00  pm--00:00 am on Sunday}
         \label{non_peak_hour_attn_144}
     \end{subfigure}
     \begin{subfigure}[b]{0.45\textwidth}
         \includegraphics[width=7cm]{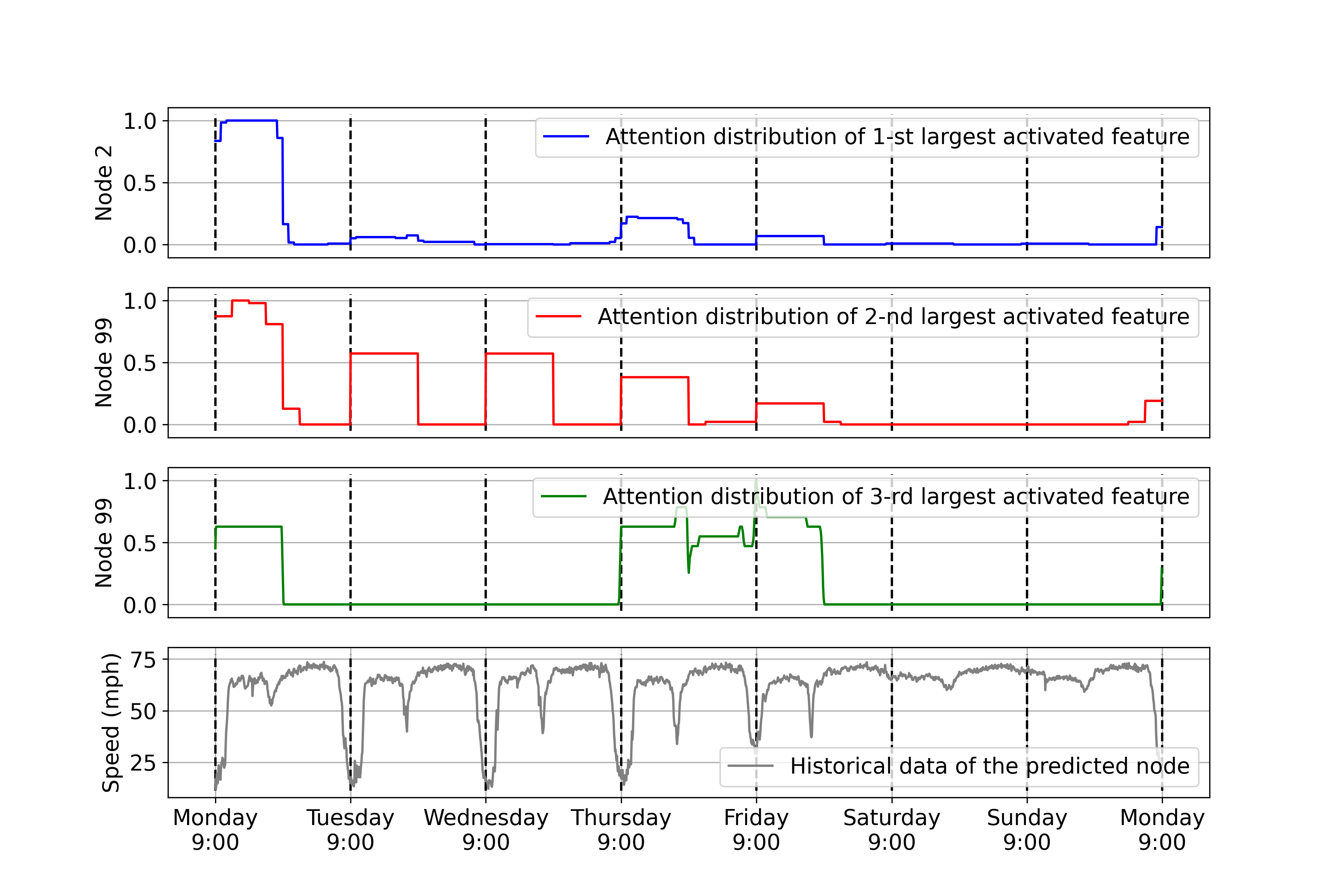}
         \caption{Prediction for 9:00  am--9:00 pm on Monday} 
         \label{peak_hour_attn_144}
     \end{subfigure}
    \caption{\footnotesize Temporal attention distribution of 5-minutes-ahead prediction for non-peak hour (top) and peak hour (bottom).}
    \label{Long_term_model_temporal_prediction}
\end{figure}

The temporal attention of the most influential node for each feature map is shown in Figure \ref{Long_term_model_temporal_prediction}. We observe that the attention not only accumulates at the same time of the last week but also on the same time of other days. This shows that our model is capable of capturing weekly and daily periodic patterns. Moreover, there is no attention on the recent traffic data, which indicates that for long-term traffic prediction the most recent traffic is not as important as for short-term traffic prediction. 

In Figure \ref{peak_hour_attn_144} we observe that to predict the peak hour of the weekday traffic, our model does not pay attention to the traffic conditions on the weekend, which shows that our model can recognize the difference between weekday traffic and weekend traffic. 

Comparing the attention distribution of short-term traffic prediction model and long-term prediction model, we conclude that spatial correlation and temporal correlation are more important for long-term prediction than for short-term prediction. In short-term prediction, spatial correlation is also utilized. Based on the results of the temporal attention distribution, however, we observe that the model actually did not extract too much information from the neighbor nodes (extracted only the information of the same time last week). This implies that short-term is a relatively easy task so that the model does not need too much information to obtain good results.

\section{Conclusion}

We studied the long-term spatial-temporal traffic forecasting problem, which is essential for intelligent traffic systems. We developed the explainable graph pyramid autoformer (X-GPA) that combines a novel pyramid autocorrelation temporal attention layers with graph spatial attention layers to learn temporal patterns from long traffic time-series and spatial dependencies, respectively. We showed that our X-GPA model achieved up to 35\% improvement in accuracy over the state-of-the-art spatial-temporal graph neural networks  developed for traffic forecasting. We used the attention-based scores from the X-GPA model to derive both spatial and temporal explanations for predictions. We show that our model can notice the congestion locations of the network spatially and capture the long-term periodic temporal pattern of the traffic system. 

Our future work will include 1) X-GPA for dynamical systems such as weather forecasting; 2) uncertainty quantification capabilities; 3) scaling for large data sets with distributed training; and 4) transfer learning.  

\clearpage

\newpage

\section{Appendix}
\label{Sec.app}

We tested the model performance for different short-term traffic forecasts (Cases 1--3). Different GNN models are built based on different inductive biases, which need different input data. Models that forecast the traffic future by simulating the evolution of the traffic system, such as HA, DCRNN, and Autoformer, use the setting of Case 1. However, models that forecast the future traffic by aggregating multiple historical traffic data, such as STGCN and ASTGCN, use the setting of Case 2. To enable a fair comparison, we  tested our proposed X-GPA performance under the same setting  of the baseline models. All experiments were conducted in PyTorch \cite{https://doi.org/10.48550/arxiv.1912.01703} on a single NVIDIA TITAN RTX 16 GB GPU.

We report the MAE of 15-min, 30-min, and 60-min ahead forecasts for Cases 1--3 in Table \ref{fig.Case_1&2_pred_performance} and Table \ref{fig.Case_3_pred_performance}. 
For Case 1, we  observe that our proposed X-GPA method obtains better forecasting accuracy than that of HA and Autoformer. DCRNN achieves slightly better accuracy than does X-GPA, where the observed difference is within 0.5 mph. For Case 2,  X-GPA achieves accuracy values that are either better than or comparable to those of  STGCN and ASTGCN. For Case 3,  our X-GPA outperforms HA, STGCN, and Autoformer. DCRNN and ASTGCN were not included in the comparison because of the high computation cost. We estimated the total training time by multiplying training time per epoch and total training epochs. DCRNN needs 100 epochs with 126 hours of training time per epoch, and ASTGCN requires 20 epochs with 75.6 hours of training time per epoch.  DCRNN was originally designed for  short-sequence input and thus is unsuitable for analyzing long-sequence input, while ASTGCN using the self-attention mechanism with $O(n^2)$ complexity is computationally expensive---the key reason that these two models need such  significant training time. Comparing the results in Cases 2 and 3, we found that by selecting only  important historical data as input, the model performance can be improved.





\begin{table}[ht]
\begin{center}
\begin{tabular}{c c c c c c c}
\multicolumn{7}{c}{MAE (in mph) comparison of Case 1 } \\ 
\hline
{} & \multicolumn{3}{c}{PEMS-bay dataset} & \multicolumn{3}{c}{Metr-LA dataset} \\
model & 15 min & 30 min & 60 min & 15 min & 30 min & 60 min\\
\hline
HA & 2.42 $\pm$ 0.00 & 2.87 $\pm$ 0.00 & 3.65 $\pm$ 0.00 & 3.47 $\pm$ 0.00 & 3.78 $\pm$ 0.00 & 4.16 $\pm$ 0.00 \\
DCRNN & \textbf{1.48 $\pm$ 0.11} & \textbf{1.81 $\pm$ 0.15} & \textbf{2.24 $\pm$ 0.21} & \textbf{2.15 $\pm$ 0.09} & \textbf{2.83 $\pm$ 0.12} & \textbf{3.68 $\pm$ 0.19}\\
Autoformer & 2.05 $\pm$ 0.13 & 2.36 $\pm$ 0.18 & 2.93 $\pm$ 0.27 & 2.55 $\pm$ 0.17 & 3.24 $\pm$ 0.19 & 3.84 $\pm$ 0.25\\
X-GPA & 1.49 $\pm$ 0.08 & 1.87 $\pm$ 0.11  & 2.23 $\pm$ 0.16 & 2.29 $\pm$ 0.11 & 2.91 $\pm$ 0.15  & 3.76 $\pm$ 0.20 \\
\hline
\end{tabular}

\medskip

\begin{tabular}{c c c c c c c}
\multicolumn{7}{c}{MAE (in mph) comparison of Case 2} \\
\hline
{} & \multicolumn{3}{c}{PEMS-bay dataset} & \multicolumn{3}{c}{Metr-LA dataset} \\
model & 15 min & 30 min & 60 min & 15 min & 30 min & 60 min\\
\hline
STGCN & 1.36 $\pm$ 0.11 & 1.81 $\pm$ 0.14 & 2.49 $\pm$ 0.20 & 2.33 $\pm$ 0.14 & 2.94 $\pm$ 0.19 & 3.79 $\pm$ 0.26 \\
ASTGCN & 1.35 $\pm$ 0.08 & 1.70 $\pm$ 0.11 & 2.06 $\pm$ 0.14 & 2.14 $\pm$ 0.13 & 2.79 $\pm$ 0.16 & 3.58 $\pm$ 0.22\\
X-GPA & \textbf{1.32 $\pm$ 0.06} & \textbf{1.62 $\pm$ 0.10} & \textbf{1.95 $\pm$ 0.15} & \textbf{2.12 $\pm$ 0.10} & \textbf{2.72 $\pm$ 0.15} & \textbf{3.50 $\pm$ 0.21} \\
\hline
\end{tabular}
\end{center}

\caption{MAE (in mph) comparison for Case 1 and Case 2. }
\label{fig.Case_1&2_pred_performance}

\end{table}

\begin{table}[ht]
\begin{center}

\begin{tabular}{c c c c c c c}
\multicolumn{7}{c}{Performance comparison of Case 3} \\
\hline
{} & \multicolumn{3}{c}{PEMS-bay dataset} & \multicolumn{3}{c}{Metr-LA dataset} \\
model & 15 min & 30 min & 60 min & 15 min & 30 min & 60 min\\
\hline
HA & 5.40 $\pm$ 0.00 & 5.40 $\pm$ 0.00 & 5.40 $\pm$ 0.00 & 15.89 $\pm$ 0.00 & 15.89 $\pm$ 0.00 & 15.89 $\pm$ 0.00\\
STGCN & 1.76 $\pm$ 0.11 & 2.31 $\pm$ 0.17 & 2.89 $\pm$ 0.23 & 2.63 $\pm$ 0.16 & 3.44 $\pm$ 0.24 & 4.19 $\pm$ 0.31 \\
Autoformer & 1.87 $\pm$ 0.10 & 2.18 $\pm$ 0.13 & 2.74 $\pm$ 0.18 & 2.81 $\pm$ 0.14 & 3.67 $\pm$ 0.20 & 4.41 $\pm$ 0.29\\
X-GPA & \textbf{1.43 $\pm$ 0.08}  & \textbf{1.79 $\pm$ 0.12} & \textbf{2.27 $\pm$ 0.19} & \textbf{2.49 $\pm$ 0.16} & \textbf{3.60 $\pm$ 0.13} & \textbf{4.07 $\pm$ 0.20}\\
\hline
\end{tabular}

\medskip

\begin{tabular}{c c c c c c c}
\multicolumn{7}{c}{Training time comparison for Case 3}\\
\hline
Model & HA & STGCN & DCRNN & ASTGCN & Autoformer & X-GPA \\
\hline
PEMS-bay dataset & 0 hours & 7.5 hours & 12600 hours & 1512 hours & 2.5 hours & 4.5 hours \\
Metr-LA dataset & 0 hours & 4.9 hours & 7052 hours & 896 hours & 1.5 hours & 3.5 hours \\
\hline
\end{tabular}
\end{center}
\caption{Model performance and efficiency comparison for Case 3. MAE is shown in mph.}
\label{fig.Case_3_pred_performance}

\end{table}

For Case 3, X-GPA and Autoformer have significantly lower training time than do the other models. Although Autoformer without the need to model the spatial dependency has lower training time than that of X-GPA, Autoformer's accuracy values are lower than those of  X-GPA. Our X-GPA model can achieve results that are superior or comparable to the state-of-the-art short-term traffic forecasting methods. 

\bibliographystyle{unsrt}
\bibliography{G_Autoformer} 

\end{document}